\documentclass[10pt,twocolumn,letterpaper]{article}

\usepackage{cvpr}              
\usepackage[accsupp]{axessibility}
\usepackage{graphicx}%
\usepackage{multirow}%
\usepackage{amsmath,amssymb,amsfonts}%
\usepackage{amsthm}%
\usepackage{mathrsfs}%
\usepackage[title]{appendix}%
\usepackage{textcomp}%
\usepackage{manyfoot}%
\usepackage{caption}
\usepackage{diagbox}
\usepackage[export]{adjustbox}
\usepackage{listings}%
\usepackage{epsfig}
\usepackage{placeins}
\usepackage{makecell}
\usepackage{float}
\usepackage{booktabs}%
\usepackage{bm}
\usepackage{setspace}
\usepackage[ruled,linesnumbered]{algorithm2e}

\usepackage[dvipsnames]{xcolor}
\newcommand{\red}[1]{{\color{red}#1}}
\newcommand{\blue}[1]{{\color{blue}#1}}

\definecolor{cvprblue}{rgb}{0.21,0.49,0.74}
\usepackage[pagebackref,breaklinks,colorlinks,citecolor=cvprblue]{hyperref}
\def\paperID{610} 
\def\confName{CVPR}
\def\confYear{2024}

\title{Selective Hourglass Mapping for Universal Image Restoration Based on Diffusion Model}

\author{Dian Zheng$^{1}$~~Xiao-Ming Wu$^1$~~Shuzhou Yang$^2$~~Jian Zhang$^2$~~Jian-Fang Hu$^1$~~Wei-Shi Zheng$^{1,3}$\thanks{Corresponding author}\\
$^1$School of Computer Science and Engineering, Sun Yat-sen University, China \\
$^2$School of Electronic and Computer Engineering, Peking University, China \\
$^3$Key Laboratory of Machine Intelligence and Advanced Computing, Ministry of Education, China \\
\tt\small \{zhengd35, wuxm65\}@mail2.sysu.edu.cn~~ wszheng@ieee.org
}

\begin{document}
{\onecolumn
\noindent \vspace{1cm}

\noindent \textbf{\huge\centering{Selective Hourglass Mapping for Universal Image Restoration Based on Diffusion Model}}

\vspace{2cm}

\noindent {\LARGE{Dian Zheng, Xiao-Ming Wu, Shuzhou Yang, \\
Jian Zhang, Jian-Fang Hu, Wei-Shi Zheng*}}
\\
\\
*Corresponding author: Wei-Shi Zheng.
\\
\\
Code: \href{https://github.com/iSEE-Laboratory/DiffUIR}{\blue{https://github.com/iSEE-Laboratory/DiffUIR}}
\\
\\
Project page: \href{https://isee-laboratory.github.io/DiffUIR/}{\blue{https://isee-laboratory.github.io/DiffUIR/}}
\\
\\
My homepage: \href{https://zhengdian1.github.io}{\blue{https://zhengdian1.github.io}}

\vspace{1cm}

\noindent {\LARGE{Accepted date: 27-Feb-2024 to  IEEE/CVF Conference on Computer Vision and Pattern Recognition}}

\vspace{1cm}

\noindent For reference of this work, please cite:

\vspace{1cm}
\noindent Dian Zheng, Xiao-Ming Wu, Shuzhou Yang, Jian Zhang, Jian-Fang Hu, and Wei-Shi Zheng. ~Selective Hourglass Mapping for Universal Image Restoration Based on Diffusion Model.~~In \emph{Proceedings of the IEEE Conference on
Computer Vision and Pattern Recognition,} 2024.

\vspace{1cm}

\noindent Bib:\\
\noindent @inproceedings\{zheng2024selective,\\
\ \ \  title     = \{Selective Hourglass Mapping for Universal Image Restoration Based on Diffusion Model\}, \\
 \ \ \   author    = \{Zheng, Dian and Wu, Xiao-Ming and Yang, Shuzhou and Zhang, Jian and Hu, Jian-Fang and Zheng, Wei-Shi\},\\
\ \ \  booktitle   = \{Proceedings of the IEEE/CVF Conference on Computer Vision and Pattern Recognition\},\\
\ \ \  year      = \{2024\}\\
\}
}
\twocolumn

\maketitle
\begin{abstract}
Universal image restoration is a practical and potential computer vision task for real-world applications. The main challenge of this task is handling the different degradation distributions at once. Existing methods mainly utilize task-specific conditions (e.g., prompt) to guide the model to learn different distributions separately, named multi-partite mapping. However, it is not suitable for universal model learning as it ignores the shared information between different tasks. In this work, we propose an advanced \textbf{selective hourglass mapping strategy} based on diffusion model, termed \textbf{DiffUIR}. Two novel considerations make our DiffUIR non-trivial. Firstly, we equip the model with strong condition guidance to obtain accurate generation direction of diffusion model (\textbf{selective}). More importantly, DiffUIR integrates a flexible shared distribution term (SDT) into the diffusion algorithm elegantly and naturally, which gradually maps different distributions into a shared one. In the reverse process, combined with SDT and strong condition guidance, DiffUIR iteratively guides the shared distribution to the task-specific distribution with high image quality (\textbf{hourglass}). Without bells and whistles, by only modifying the mapping strategy, we achieve state-of-the-art performance on five image restoration tasks, 22 benchmarks in the universal setting and zero-shot generalization setting. Surprisingly, by only using a lightweight model (only 0.89M), we could achieve outstanding performance. The source code and pre-trained models are available at \textcolor{magenta}{https://github.com/iSEE-Laboratory/DiffUIR} 
\end{abstract}


\begin{figure}[t]
    \centering
    \includegraphics[width=1.\linewidth]{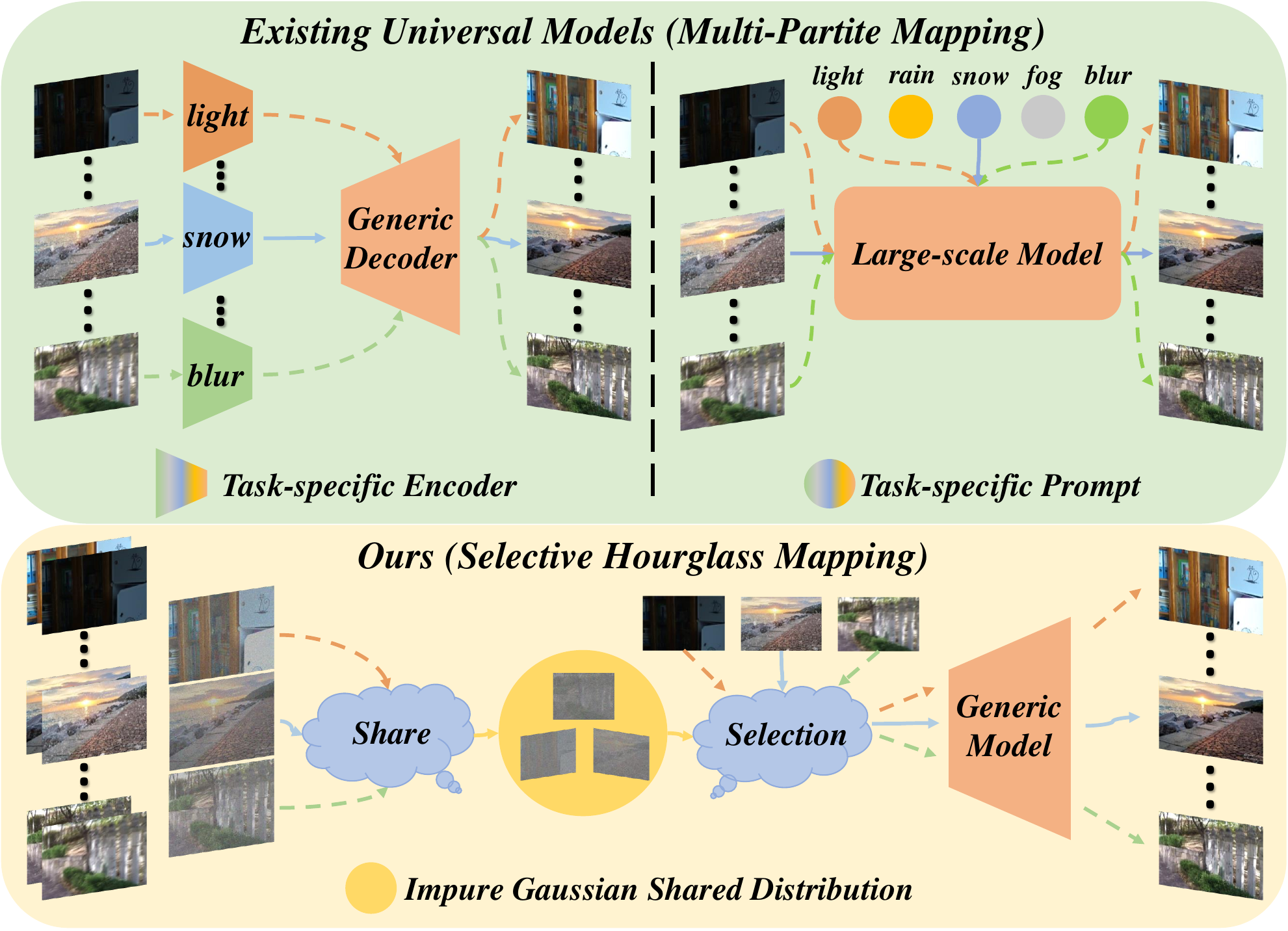}
    \caption{An illustration of existing universal image restoration methods compared with our DiffUIR, existing methods mainly design task-specific modules to handle different distributions, which force the generic model (tangerine module) to learn different distributions at once, termed multi-partite mapping. In contrast, the proposed DiffUIR maps the different distributions to one shared distribution (\ie, note that it is not the pure Gaussian distribution) while maintaining strong condition guidance. In this way, DiffUIR enables the generic model to only learn one shared distribution and guides the shared distribution to a task-specific distribution, termed selective hourglass mapping. Zoom in for best view.}
    \label{fig:frame_all}
    \vspace{-4mm}
\end{figure}

\section{Introduction}
\label{sec:intro}
Universal image restoration aims to handle different image restoration tasks on a single model, which has a wide range of applications in real-world robotics navigation~\cite{2011robot} and autonomous driving~\cite{2015deepdriving}.
\par
The main challenge of handling different image restoration tasks by a single model is learning various distributions simultaneously. Existing universal image restoration methods~\cite{multi-encoder, Painter, promptir, Prores, daclip} mainly utilize the multi-encoder architecture or prompt the large-scale models as shown in Fig.~\ref{fig:frame_all}. In this way, according to~\cite{single_learn}, they will learn separate mappings in one model by guiding the model to learn different distributions separately when meeting specific conditions (\ie, multi-partite mapping). Despite the strong condition guidance maintaining certain image quality, they ignore the fact that different tasks may share information that has the potential to complement and enhance the performance of the single task. For instance, in deraining dataset, rainy and foggy weather invariably occur simultaneously. Attempting to learn these two degradation types independently might fall short in addressing this scenario.
\par
In this work, we aim to capture the shared information between different tasks for better universal image restoration learning. We replace the multi-partite mapping strategy with a novel selective hourglass mapping strategy based on conditional diffusion model, termed DiffUIR. Two novel designs make our DiffUIR non-trivial. Firstly, inspired by the RDDM~\cite{RDDM}, we explicitly fuse the condition (\ie, degraded images) into the diffusing algorithm of the diffusion model and extensively concatenate the condition with the diffusing target. In this way, DiffUIR is equipped with the ability of strong condition guidance which is similar to multi-partite mapping methods. Secondly, to achieve shared distribution mapping, we integrate a shared distribution term, named SDT into the diffusion algorithm elegantly and naturally, which gradually adjusts the weight of the condition in the algorithm. By modeling the two problems, in the forward diffusing process, DiffUIR gradually reduces the weight of the condition and the various distributions will approach one shared distribution, enabling the model to capture the shared information between tasks. Notably, we map the different distributions to an impure  Gaussian distribution with slight condition left, as mentioned in~\cite{2023inver,bridge,RDDM}, the pure Gaussian noise contains no task information, which is not conducive to good generation quality. In the reverse process, by the guidance of the strong condition and SDT, DiffUIR will gradually recover from the shared distribution into the task-specific distribution.
\par
By only modifying the mapping strategies, without bells and whistles we outperform all of the existing universal methods by a large margin in five image restoration benchmarks. Notably, we only use the model with 36.26M parameters, which is at least 5 times less than the existing large-scale model based universal methods but with higher performance. Additionally, to meet the demand of real-world applications, we propose several light versions of our DiffUIR, the tiny version DiffUIR-T comprises only 0.89M parameters, yet it exhibits outstanding performance. To further validate the ability of our universal model, we do zero-shot generalization experiment on known task and unknown task settings, which also achieve state-of-the-art performance compared with other universal methods.
\par
In summary, our main contributions are as follows:   \par
$(1)$ A novel selective hourglass mapping method, DiffUIR is proposed, which can freely convert various distributions to a shared one and enable the model to learn shared information between different tasks. Moreover, equipped with strong conditions, DiffUIR guides the shared distribution to a task-specific distribution with high image quality.
\par
$(2)$ We empirically validate that our distribution mapping strategy is a better solution for the universal image restoration task. By only changing the mapping strategy, we even outperform the universal image restoration methods based on large-scale models with only $1/5$ parameters.\par
$(3)$ Our DiffUIR meets the demand of real-world scenes. We outperform other universal methods in zero-shot generalization setting. Our tiny version DiffUIR-T comprised only 0.89M parameters with outstanding performance.

\vspace{-2mm}
\section{Related Work}
\vspace{-1mm}
\label{sec:relate}
\subsection{Image Restoration.}
\vspace{-1mm}
Image restoration aims to recover a clean image from its degradation counterpart, which is a fundamental and significant computer vision field and contains various tasks, such as deraining, desnowing, low-light enhancement, debluring and dehazing, \etc Existing works~\cite{he2010single, yangshuzhou, drs, ir_sde, mirnet, DehezeFormer, Restoreformer, WeatherDiff, RDDM} mainly focus on resolving one specific task with unique model designs. Although these methods achieve great performance success, they ignore the fact that in real-world applications, people prefer one model that can handle all of the degradation types. Recently some pioneers have studied on universal image restoration model and made some progress. AirNet~\cite{airnet} uses a module to map the different distributions into one shared distribution constrained by contrastive learning, which is hard for training and the performance is limited. IDR~\cite{IDR} observes that different degradation types can be divided by singular value decomposition and the clean image could be recomputed by reformulating the singular value and vector. Painter~\cite{Painter}, ProRes~\cite{Prores} and DA-CLIP~\cite{daclip} aim to incorporate the full potential of large-scale models by prompt learning. Although they use the prior knowledge of the large-scale model, due to utilizing multi-partite mapping strategy, they only perform limited performance and require a vast number of parameters.
\par
In this work, we propose selective hourglass mapping strategy based on conditional diffusion model, equipped the model with the ability of shared distribution mapping and strong condition guidance at once. Due to these capabilities, we achieve outstanding results without relying on a complex training pipeline or large-scale models or pre-training.


\subsection{Diffusion Model}
\vspace{-1mm}
Diffusion model is a new kind of generative model which fits the distribution of the empirical data from standard Gaussian distribution based on the Markov Chain. We show the theoretical and implementation details in $\textit{Appendix}$ A. Benefiting from the strong mathematical basis and great generative quality, diffusion model is widely used in various dense computer vision tasks, such as image generation~\cite{score_image_generation, i2i, ddpm, ddim,stable_diffusion}, image editing~\cite{diffedit, blendededit, freedom}, image segmentation~\cite{baranchuk2021label, panoptic}, depth estimation~\cite{diffustereo, diffuvolume}, \etc. As a part of dense estimation tasks, many researchers apply the diffusion model to image restoration. RainDiffusion~\cite{raindiffusion} fuses the cycle framework into the conditional diffusion model and performs well in unsupervised setting, DDNM~\cite{ddnm} constructs an elegant identity equation, which naturally adds condition into the reverse process of diffusion model without any extra training, working greatly on linear image restoration tasks. RDDM~\cite{RDDM} changes the diffusion direction from the target domain to the input domain, which naturally integrates the condition (\ie, degraded image) into the forward process and achieves impressive performance on several image restoration benchmarks.
\par
The approaches above present some intriguing modifications concerning conditional constraints, achieving strong condition guidance. However, they are all not suitable for universal image restoration learning. The diffusing endpoint of the standard conditional diffusion model is the standard Gaussian noise without any task-specific context information, as mentioned in~\cite{RDDM,2023inver,bridge}, it suffers from poor recovering quality because the condition is added in a mediated way (\ie, concatenation); RDDM~\cite{RDDM} explicitly fuses the condition into the diffusing algorithm, achieving high image quality. However, the problem of multi-partite mapping occurs as the endpoints of different tasks are distinguishable and pertain to different distributions. In this work, we address the shortcomings of existing conditional diffusion models and achieve shared distribution mapping and strong condition modeling at the same time.

\section{DiffUIR}
In this section, we first explore a suitable condition mechanism of diffusion model inspired by RDDM, then we introduce our selective hourglass mapping strategy which equips the abilities of shared distribution mapping and strong condition guidance for better universal learning. 

\subsection{Revist the condition mechanism of RDDM}
\vspace{-1mm}
RDDM~\cite{RDDM} follows standard T-step diffusion model~\cite{ddim,ddpm} which contains a forward process and a reverse process. In the forward process, the one-step noising could be written as Markov Chain:
\vspace{-2mm}
\begin{equation}
\label{eqn:rddm_f1}
    q(I_t|I_{t-1}, I_{res}) = \mathcal{N}(I_t; I_{t-1}+\alpha_tI_{res}, \beta^2_t\textbf{I}),
\vspace{-2mm}
\end{equation}
where $I_t$ is the diffusing result in timestep \textit{t}, $I_{res}$ is the residual of degraded image $I_{in}$ and clean image $I_0$: $I_{res}$ = $I_{in}$ - $I_{0}$, $\alpha_t$ and $\beta_t$ is the noise coefficient of $I_{res}$ and Gaussian noise respectively. They change the noising objective from the $I_0$ (\ie, used in previous image restoration diffusion methods~\cite{raindiffusion, WeatherDiff}) to $I_{res}$ following the residual learning~\cite{resnet}. By the property of Markov Chain and reparameterization~\cite{vae, inro_vae} technology, the one-step noising distribution could be extended to any step noising form as follows:
\vspace{-2mm}
\begin{equation}
\label{eqn:rddm_f2}
    q(I_t|I_0, I_{res}) = \mathcal{N}(I_t; I_0+\overline{\alpha}_tI_{res}, \overline{\beta}_t\textbf{I}),
\vspace{-2mm}
\end{equation}
where $\overline{\alpha}_t=\sum_{i=1}^t\alpha_i$, $\overline{\beta}_t=\sqrt{\sum_{i=1}^t\beta^2_i}$. Notably when $t \rightarrow T$, $\overline{\alpha}_T = 1$ and the formula could be written as $I_T = I_{in} + \overline{\beta}_T\epsilon$. It shows that the endpoint is only related to the degraded image and the added noise, naturally adding the condition to the model training. In the reverse process, RDDM uses the $q(I_{t-1}|I_t,I_0^{\theta},I_{res}^{\theta})$ to simulate the true generation distribution $p_{\theta}(I_{t-1}|I_t)$, and it could also be written as Markov Chain:
\vspace{-2mm}
\begin{equation} 
\label{eqn:rddm_r}
    p_{\theta}(I_{t-1}|I_t) = \mathcal{N}(I_{t-1}; I_0^{\theta}+\overline{\alpha}_{t-1}I_{res}^{\theta}+\overline{\beta}_{t-1}\epsilon^{\theta}, 
 \textbf{0}\cdot\textbf{I}),
\vspace{-2mm}
\end{equation}
where the terms with $\theta$ mean that they are obtained based on the model output, $\textbf{0}$ occurs as they use the deterministic implicit sampling equation following~\cite{ddim}.
\par
As the endpoint of the RDDM contains information of the condition (\ie, degraded image), it is a great condition mechanism of diffusion model, we call it explicit condition, what's more, they extensively concatenate the condition with the diffusing target, benefiting of better image quality, named implicit condition. However, the strong condition mechanism in RDDM is not suitable for universal training as the condition always exists, which means they force the model to learn different degraded distributions separately, causing multi-partite mapping and the shared information between different tasks could not be captured. 
\subsection{Selective Hourglass Mapping}
\vspace{-1mm}
\label{sec:forward}
The goal of our method is to achieve strong condition guidance and shared distribution mapping at the same time. We adopt the condition mechanism of RDDM~\cite{RDDM} and integrate a shared distribution term (SDT) into the diffusion algorithm that achieves the synergistic effect between two components. We show our variant diffusion process below.
\par
\noindent\textbf{Distribution Approaching Forward Process.}
In the forward process, as we adopt the condition mechanism of RDDM~\cite{RDDM}, the one-step diffusion process is: $I_t = I_{t-1}+\alpha_tI_{res}+ \beta_t\epsilon_{t-1}.$ To further achieve shared distribution mapping, we modify the forward process as follows:
\vspace{-2mm}
\begin{equation}
\begin{aligned}
\label{eqn:our_f1}
    I_t &= I_{t-1}+\alpha_tI_{res}+ \beta_t\epsilon_{t-1}-\delta_tI_{in} \\
        &= I_{t-2}+(\alpha_t+\alpha_{t-1})I_{res}+ \sqrt{\beta^2_t+\beta^2_{t-1}}\epsilon_{t-2}\\
        &\quad-(\delta_t+\delta_{t-1})I_{in} \\
        &= I_{t-3} \dots \\
        &= I_0+\overline{\alpha}_tI_{res}+\overline{\beta}_t\epsilon-\overline{\delta}_tI_{in},
\end{aligned}
\vspace{-2mm}
\end{equation}
where  $\delta_tI_{in}$ is the SDT, $\delta$ is the shared distribution coefficient and $\overline{\delta}_t=\sum_{i=1}^t\delta_i$. We set the value of $\overline{\delta}_t$ from 0 to 0.9, which will gradually reduce the influence of the condition. When $t \rightarrow T$, $\overline{\alpha}_T = 1$, and the formula could be rewritten as: $I_T=(1-\overline{\delta}_T)I_{in}+\overline{\beta}_T\epsilon = 0.1I_{in} + \overline{\beta}_T\epsilon$, which approaches an impure Gaussian distribution (we further validate it in experiment). Note that we adopt the progressive approaching strategy to fit the diffusing process of the diffusion model naturally.
\par
\begin{algorithm}[t]
\setstretch{0.95}
	\caption{Training.}
	\label{alg:algorithm1}
        \KwIn{Clean image: $I_0$; \\
        \quad\quad\quad Degraded image: $I_{in}$; \\
        \quad\quad\quad Residual map: $I_{res} = I_{in} - I_{0}$.}
        \Repeat{\textnormal{converged}}{
        $I_0 \sim q(I_0)$; \\
        $t \sim Uniform({1,\dots,T})$; \\
        $\epsilon \sim \mathcal{N}(\bm{0},\bm{I})$; \\
        $I_t = I_0+\overline{\alpha}_tI_{res}+\overline{\beta}_t\epsilon-\overline{\delta}_tI_{in}$; \\
        Take the gradient descent step on \\
        \quad\quad $\nabla_{\theta} \Vert I_{res} - I_{res}^{\theta}(I_t, I_{in}, t) \Vert_1 $;
	}
\end{algorithm}
\noindent\textbf{Distribution Diffusing Reverse Process.}
In the reverse process (inference stage), our goal is to recover the sample from the shared distribution (\ie, $I_T=(1-\overline{\delta}_T)I_{in}+\overline{\beta}_T\epsilon$) to the task-specific distribution. Following the DDPM~\cite{ddpm}, we use the $q(I_{t-1}|I_t,I_{in},I_0^{\theta},I_{res}^{\theta})$ to simulate the distribution of $p_{\theta}(I_{t-1}|I_t)$ and based on the Bayes’ theorem, we could calculate it as follows:
\vspace{-3mm}
\begin{equation}
\begin{aligned}
\label{eqn:bayes}
    &\ p_{\theta}(I_{t-1}|I_t) \rightarrow q(I_{t-1}|I_t,I_{in},I_0^{\theta},I_{res}^{\theta}) & \\
    &= q(I_t|I_{t-1},I_{in},I_{res}^{\theta})\frac{q(I_{t-1}|I_0^{\theta},I_{res}^{\theta},I_{in})}{q(I_t|I_0^{\theta},I_{res}^{\theta},I_{in})}  \\
    &\propto exp\left[-\frac{1}{2}((\frac{\overline{\beta}_t^2}{\beta_t^2\overline{\beta}_{t-1}^2})I_{t-1}^2-2(\frac{I_t+\delta_t I_{in}-\alpha_tI_{res}^{\theta}}{\beta_t^2} \right.\\
    &\left.+\frac{I_0^{\theta}+\overline{\alpha}_{t-1}I_{res}^{\theta}-\overline{\delta}_{t-1}I_{in}}{\overline{\beta}_{t-1}^2})I_{t-1}+C(I_t,I_0^{\theta},I_{res}^{\theta},I_{in}))\right].
\end{aligned}
\vspace{-1mm}
\end{equation}
As the goal of the formulation is to obtain the distribution of the $I_{t-1}$, we simplify and rearrange it into the form about the $I_{t-1}$ and $C(I_t,I_0^{\theta},I_{res}^{\theta},I_{in})$ is the term unrelated with it. Then we calculate the mean $\mu_{\theta}(I_t, t)$ and variance $\sigma_{\theta}(I_t, t)$ of the distribution $p_{\theta}(I_{t-1}|I_t)$ based on Eq. (\ref{eqn:bayes}):
\vspace{-3mm}
\begin{equation}
\begin{aligned}
\label{eqn:our_mean}
    &\mu_{\theta}(I_t, t) = I_t - \alpha_t I_{res}^{\theta} + \delta_t I_{in} - \frac{\beta^2_t}{\overline{\beta}_t}\epsilon^{\theta} \\
    &\sigma_{\theta}(I_t, t) = \frac{\beta^2_t \overline{\beta}^2_{t-1}}{\overline{\beta}^2_t},
\end{aligned}
\vspace{-3mm}
\end{equation}
where $I_{res}^{\theta}$ is predicted by the model and $\epsilon^{\theta}$ is obtained by the $I_{res}^{\theta}$. Based on the reparameterization~\cite{vae, inro_vae} technology, if we use the sampling strategy from the DDPM~\cite{ddpm},  $I_{t-1}$ could be calculated as follows:
\vspace{-3mm}
\begin{equation}
\label{eqn:our_ddpm}
    I_{t-1} = I_t - \alpha_t I_{res}^{\theta} + \delta_t I_{in} - \frac{\beta^2_t}{\overline{\beta}_t}\epsilon^{\theta} + \frac{\beta_t \overline{\beta}_{t-1}}{\overline{\beta}_t}\epsilon_*,
\vspace{-3mm}
\end{equation}
where $\epsilon_*$ is the random Gaussian noise. In this paper, to accelerate the sampling speed, we use the sampling strategy of DDIM~\cite{ddim} and $I_{t-1}$ is calculated by:
\vspace{-2mm}
\begin{equation}
\label{eqn:oue_ddim}
    I_{t-1} = I_t - \alpha_t I_{res}^{\theta} + \delta_t I_{in}.
    \vspace{-2mm}
\end{equation}
Based on the Eq. (\ref{eqn:oue_ddim}), we could iteratively recover the sample from the $I_T$ to $I_{T-k}, I_{T-2k}, \dots, I_k, I_0$ where k means skip steps followed DDIM~\cite{ddim}. The complete sampling algorithm is shown in Algorithm~\ref{alg:algorithm2}.
\par
\textbf{Note} that we only present the most important steps here and the full derivation please refer to \textit{Appendix} C.

\begin{algorithm}[t]
\setstretch{0.95}
	\caption{Sampling.}
	\label{alg:algorithm2}
        \KwIn{Degraded image: $I_{in}$.}
        $\epsilon \sim \mathcal{N}(\bm{0},\bm{I})$; \\
        $I_T = (1-\overline{\delta}_T)I_{in} + \overline{\beta}_T\epsilon$; \\
        \For {$t=T,\dots,1 $}{
        if $t > 1$ \\
        
        \quad $I_{t-1} = I_t - \alpha_t I_{res}^{\theta}(I_t, I_{in}, t) + \delta_t I_{in}$; \\
        else \\
        \quad $I_{t-1} = I_{in} - I_{res}^{\theta}(I_t, I_{in}, t)$;\\
        }
        \Return {$I_0$}
\end{algorithm}

\begin{figure*}[t]
    \centering
    \includegraphics[width=1.\linewidth]{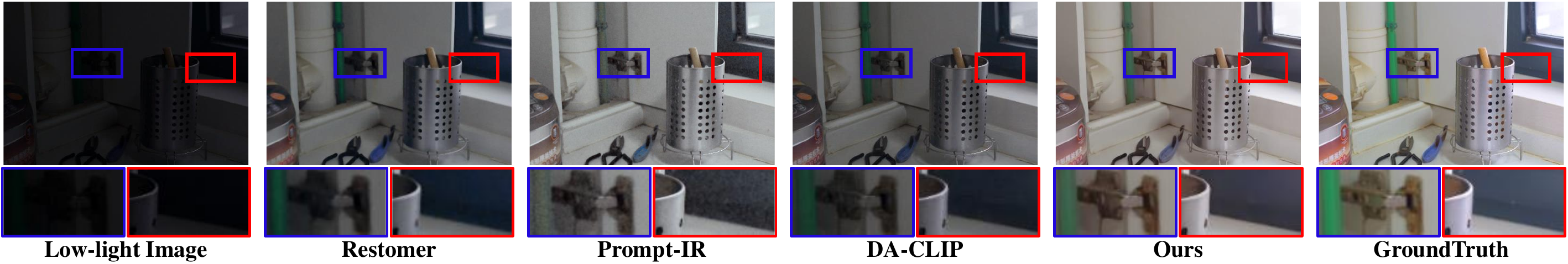}
    \vspace{-7mm}
    \caption{Visualization comparison with state-of-the-art methods on low-light enhancement. Zoom in for best view.}
    \vspace{-5mm}
    \label{fig_light}
\end{figure*}

\noindent\textbf{Universal Training Objective.}
We achieve universal learning by modifying the condition mechanism of diffusion model into explicit condition (injected into the diffusion algorithm) and implicit condition (concatenation) for strong condition guidance; adding SDT into the diffusion algorithm for shared distribution mapping. Although it is different from standard diffusion models~\cite{ddpm, ddim}, as we both approach the different degraded distributions to one shared distribution, we reference DDPM~\cite{ddpm} and conduct the meticulous derivation of the training objective as follows:
\vspace{-3mm}
\begin{equation}
\label{eqn:objective}
    \mathcal{L}(\theta) = D_{KL}(q(I_{t-1}|I_t,I_0,I_{res},I_{in}) || p_{\theta}(I_{t-1}|I_t)).
    \vspace{-1mm}
\end{equation}
As mentioned in VAE~\cite{vae}, the Kullback-Leibler divergence of two Gaussian distributions could be simplified to the difference of their mean, the function is transformed as:
\vspace{-3mm}
\begin{equation}
\begin{aligned}
\label{eqn:mean_kl}
    \mathcal{L}(\theta) &= \mathbb{E}_{q(I_t|I_0)}\left[\Vert \mu(I_t,I_0) - \mu_{\theta}(I_t,t) \Vert^2\right] \\
                        &= \mathbb{E}_{t,\epsilon,I_{res}}\left[\Vert I_t - \alpha_t I_{res} + \delta_t I_{in} - \frac{\beta^2_t}{\overline{\beta}_t}\epsilon - \right.\\
                        & \left.\quad (I_t - \alpha_t I_{res}^{\theta} + \delta_t I_{in} - \frac{\beta^2_t}{\overline{\beta}_t}\epsilon^{\theta}) \Vert^2\right] \\
                        &= \mathbb{E}_{t,\epsilon,I_{res}}\left[\Vert \alpha_t(I_{res}^{\theta} - I_{res})+\frac{\beta^2_t}{\overline{\beta}_t}(\epsilon^{\theta}-\epsilon) \Vert^2\right], \\
\end{aligned}
\end{equation}
where $I_{res}$, $\epsilon$ mean established value in the forward process and $I_{res}^{\theta}$, $\epsilon^{\theta}$ mean predicted result in the reverse process. \par
Referring to the official code of DDPM~\cite{ddpm}, predicting the noise or the input is essentially equivalent, so we directly use the model to predict the residual $I_{res}^{\theta}$, the $\epsilon^{\theta}$ could be derivated by it. According to ~\cite{pdpp,diffuvolume,diffusiondet}, when predicting the input, $L_1$ loss performs better than $L_2$ loss. Based on the experimental and theoretical basis above, our final loss function is simplified as follows:
\vspace{-2mm}
\begin{equation}
\label{eqn:loss}
    \mathcal{L}(\theta)_{simple} = \mathbb{E}_{t,I_t,I_{res}}\left[\Vert I_{res} - I^{\theta}_{res}(I_t,t) \Vert_1\right].
    \vspace{-3mm}
\end{equation}

The full training algorithm is shown in Algorithm~\ref{alg:algorithm1}.

\section{Experiments}
\subsection{Datasets and Evaluation Metrics}
\vspace{-1mm}
We evaluate the proposed DiffUIR on five image restoration tasks. We select the most widely used datasets for each task as follows, the summary table is in $\textit{Appendix}$ B:     \par
\noindent\textbf{Image deraining.} We use the merged datasets mentioned in~\cite{mspfn,Restoreformer}, which covers a wide range of different rain streaks and rain density. The datasets contain 13712 data for training and 4298 data for testing, providing a robust platform to evaluate the performance of deraining methods. We further perform zero-shot generalization on real-world dataset Practical~\cite{rain100} which has no ground truth.
\par
\noindent\textbf{Low-light enhancement.} We use the famous LOL~\cite{lol} datasets as the benchmark with 485 training pairs and 15 testing pairs, which consist of a large number of indoor and outdoor scenes with different levels of light and noise. We further generalize to the widely used real-world datasets MEF~\cite{mef}, NPE~\cite{npe_loe} and DICM~\cite{dice} without ground truth.
\par
\noindent\textbf{Image desnowing.} The Snow100K~\cite{snow100} is used for desnowing benchmark with 50000 training data, 50000 testing data and 1329 real-world snow images. In this work, we test on Snow100K-S and Snow100K-L which means light and hazy snowflake sizes and further perform zero-shot generalization on real-world snow pairs.
\par
\noindent\textbf{Image dehazing.} We use RESIDE~\cite{reside} datasets as the dehazing benchmark. It is a widely used synthetic fog dataset with different hazing levels of fog and scenes. As the real-world fog condition is outdoor, we only train and test on the outdoor part with 313950 and 500 data respectively.
\par
\noindent\textbf{Image deblurring.} We use the GoPro~\cite{gopro} dataset as the deblurring benchmark with 2103 training pairs and 1111 testing pairs. It contains various levels of blur obtained by averaging the clear images captured in very short intervals. To further validate the power of our model, we perform zero-shot generalization on HIDE~\cite{hide}, RealBlur-J~\cite{realblur} and RealBlur-R~\cite{realblur} with 2025, 980, 980 pairs respectively.
\par

\begin{table*}[t]
    \centering
    \caption{Comparison of our method with other task-specific and universal image restoration approaches in five image restoration tasks. $\dagger$ means reimplementing in our datasets for fair comparison. The FLOPS is calculated in the inference stage with 256 $\times$ 256 resolution. The best results of task-specific models and universal models are shown in \blue{blue} and \red{red} respectively.}
    \vspace{-2mm}
    \resizebox{2.1\columnwidth}{!}{
        \begin{tabular}{ll|c|cc|cc|cc|cc|cc|cc}
        \toprule[0.15em]
        \multicolumn{2}{l|}{\multirow{2}{*}{\textbf{Method}}} & \multirow{2}{*}{\textbf{Year}} &
        \multicolumn{2}{c|}{\textbf{Deraining} $(5sets)$} & \multicolumn{2}{c|}{\textbf{Enhancement}} & \multicolumn{2}{c|}{\textbf{Desnowing} $(2sets)$} & \multicolumn{2}{c|}{\textbf{Dehazing}} & \multicolumn{2}{c|}{\textbf{Deblurring}} & \multicolumn{2}{c}{\textbf{Complexity}}\\
        \multicolumn{2}{c|}{} & & PSNR $\uparrow$ & SSIM $\uparrow$ & PSNR $\uparrow$ & SSIM $\uparrow$ & PSNR $\uparrow$ & SSIM $\uparrow$ & PSNR $\uparrow$ & SSIM $\uparrow$  & PSNR $\uparrow$ & SSIM $\uparrow$ & Params (M) & FLOPS (G)\\
        \midrule[0.15em]
        \multicolumn{14}{l}{\textbf{Task-specific Method}} \\
        \midrule[0.1em]
        \multicolumn{2}{l|}{SwinIR~\cite{swinir}} & 2021 & - & - & 17.81 & 0.723 & - & - & 21.5 & 0.891 & 24.52 & 0.773 & \blue{0.9} & 752.13 \\
        \multicolumn{2}{l|}{MIRNet-v2~\cite{mirnet}} & 2022 & - & - & \blue{24.74} & 0.851 & - & - & 24.03 & 0.927 & 26.30 & 0.799 & 5.9 & 140.92 \\
        \multicolumn{2}{l|}{DehazeFormer~\cite{DehezeFormer}} & 2022 & - & - & - & -  & - & - & \blue{34.29} & 0.983 & - & - & 4.63 & 48.64 \\
        \multicolumn{2}{l|}{Restomer~\cite{Restoreformer}} & 2022 & \blue{33.96} & \blue{0.935} & 20.41 & 0.806 & - & - & 30.87 & 0.969 & \blue{32.92} & \blue{0.961} & 26.12 & 141 \\
        \multicolumn{2}{l|}{MAXIM~\cite{maxim}} & 2022 & 33.24 & 0.933 & 23.43 & 0.863 & - & - & 34.19 & \blue{0.985} & 32.86 & 0.940 & 14.1 & 216 \\
        \multicolumn{2}{l|}{DRSFormer~\cite{drs}} & 2023 & 33.15 & 0.927 & - & - & - & - & - & - & - & - & 33.7 & 242.9 \\
        \multicolumn{2}{l|}{IR-SDE~\cite{ir_sde}} & 2023 & - & - & 20.45 & 0.787 & - & - & - & - & 30.70 & 0.901 & 34.2 & 98.3\\
        \multicolumn{2}{l|}{WeatherDiff~\cite{WeatherDiff}} & 2023 & - & - & - & - & \blue{33.51} & \blue{0.939} & - & - & - & - & 82.96 & -\\
        \multicolumn{2}{l|}{RDDM~\cite{RDDM}$\dagger$} & 2023 & 30.74 & 0.903 & 23.22 & \blue{0.899} & 32.55 & 0.927 & 30.78 & 0.953 & 29.53 & 0.876 & 36.26 & \blue{9.88} \\
        \midrule[0.1em]
        \multicolumn{14}{l}{\textbf{Universal Method}}\\
        \midrule[0.1em]
        \multicolumn{2}{l|}{Restomer$\dagger$} & - & 27.10 & 0.843 & 17.63 & 0.542  & 28.61 & 0.876 & 22.79 & 0.706 & 26.36 & 0.814 & 26.12 & 141 \\
        \multicolumn{2}{l|}{AirNet~\cite{airnet}$\dagger$} & 2022  & 24.87 & 0.773 & 14.83 & 0.767 & 27.63 & 0.860 & 25.47 & 0.923 & 26.92 & 0.811 & 8.93 & 30.13 \\
        \multicolumn{2}{l|}{Painter~\cite{Painter}} & 2022 & 29.49 & 0.868 & 22.40 & 0.872 & - & - & - & - & - & - & 307 & 248.9 \\
        \multicolumn{2}{l|}{IDR~\cite{IDR}} & 2023 & - & - & 21.34 & 0.826 & - & - & 25.24 & 0.943 & 27.87 & 0.846 & 15.34 & - \\
        \multicolumn{2}{l|}{ProRes~\cite{Prores}} & 2023 & 30.67 & 0.891 & 22.73 & 0.877 & - & - & - & - & 27.53 & 0.851 & 307 & 248.9 \\
        \multicolumn{2}{l|}{Prompt-IR~\cite{promptir}$\dagger$} & 2023 & 29.56 & 0.888 & 22.89 & 0.847 & 31.98 & 0.924 & 32.02 & 0.952 & 27.21 & 0.817 & 35.59 & 15.81 \\
        \multicolumn{2}{l|}{DA-CLIP~\cite{daclip}$\dagger$} & 2023 & 28.96 & 0.853 & 24.17 & 0.882 & 30.80 & 0.888 & 31.39 & \red{0.983} & 25.39 & 0.805 & 174.1 & 118.5 \\
        \midrule[0.1em]
        \multicolumn{2}{l|}{\textbf{DiffUIR-T(ours)}} & -  & 28.60 & 0.876 & 23.01 & 0.880 & 30.34 & 0.900 & 30.83 & 0.953 & 27.25 & 0.817 & \red{0.89} & \red{1.72} \\
        \multicolumn{2}{l|}{\textbf{DiffUIR-S(ours)}} & -  & 30.25 & 0.893 & 23.52 & 0.895 & 31.45 & 0.915 & 31.83 & 0.954 & 27.79 & 0.830 & 3.27 & 2.40 \\
        \multicolumn{2}{l|}{\textbf{DiffUIR-B(ours)}} & -  & 30.56 & 0.901 & 25.06 & 0.900 & 32.37 & 0.924 & 32.70 & 0.956 & 28.54 & 0.851 & 12.41 & 7.19 \\
        \multicolumn{2}{l|}{\textbf{DiffUIR-L(ours)}} & -  & \red{31.03} & \red{0.904} & \red{25.12} & \red{0.907} & \red{32.65} & \red{0.927} & \red{32.94} & 0.956 & \red{29.17} & \red{0.864} & 36.26 & 9.88 \\
        \bottomrule[0.1em]
    \end{tabular}{}
    }
    \vspace{-2mm}
    \label{tab:all}
\end{table*}

\begin{figure*}[t]
    \centering
    \includegraphics[width=1.\linewidth]{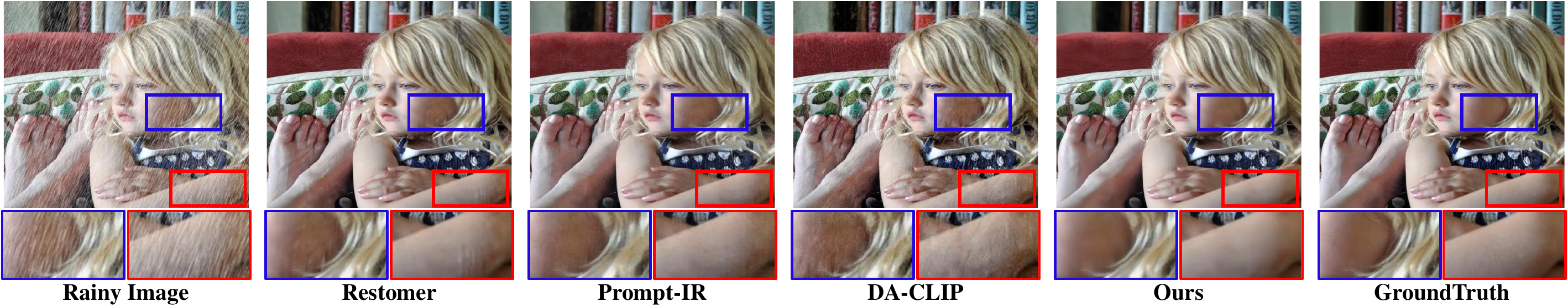}
    \vspace{-7mm}
    \caption{Visualization comparison with state-of-the-art methods on deraining. Zoom in for best view.}
    \vspace{-6mm}
    \label{fig_rain}
\end{figure*}

\begin{figure*}[t]
    \centering
    \includegraphics[width=1.\linewidth]{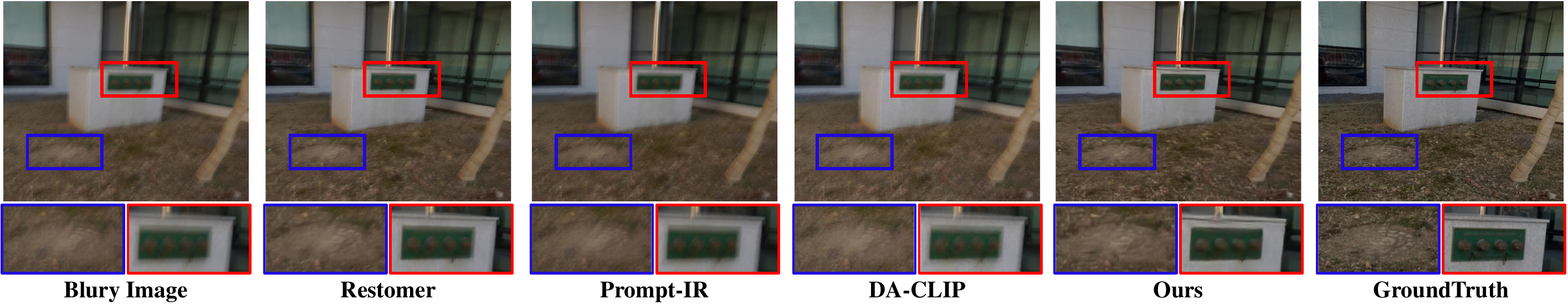}
    \vspace{-7mm}
    \caption{Visualization comparison with state-of-the-art methods on debluring. Zoom in for best view.}
    \vspace{-3mm}
    \label{fig_blur}
\end{figure*}

The benchmarks are evaluated on peak signal-to-noise ratio (PSNR)~\cite{psnr}, structural similarity (SSIM)~\cite{ssim}, natural image quality evaluator (NIQE)~\cite{niqe}, lightness order error (LOE)~\cite{npe_loe}, integrated local NIQE (IL -NIQE)~\cite{ilniqe} and learned perceptual image patch similarity (LPIPS)~\cite{lpips} metrics. Unless mentioned otherwise, PSNR and SSIM are computed using the Y channel in YCbCr color space in a way similar to existing methods~\cite{jiang2020multi, zamir2021multi, purohit2021spatially, Restoreformer, RDDM}.

\subsection{Implementation Details}
\vspace{-1mm}
DiffUIR is trained from scratch using one RTX 4090 GPU with PyTorch~\cite{paszke2019pytorch} environment for 76h. During training, We use the Adam~\cite{kingma2014adam} optimizer and L1 loss for 300k iterations with the initial learning rate 8e-5. We set the batch sizes as 10. As the data size varies greatly from task to task, we set the weight of each task in one batch as 0.4 for dehazing, 0.1 for low-light, 0.2 for deraining, 0.2 for desnowing and 0.1 for debluring. For data augmentation, we use horizontal and vertical flips for all and preprocess the low-light data with histogram equalization. We random crop 256$\times$256 patch from the original image as network input after data augmentation for training. During inference, we \textbf{test in full resolution} and use 3 timesteps for all the tasks. We use U-Net~\cite{ronneberger2015unet} architecture and the hyper-parameters of our different versions are as follows:
\begin{itemize}
  \item DiffUIR-T: $C = 32$, channel multiplier = $\{1,1,1,1\}$
  \item DiffUIR-S: $C = 32$, channel multiplier = $\{1,2,2,4\}$
  \item DiffUIR-B: $C = 64$, channel multiplier = $\{1,2,2,4\}$
  \item DiffUIR-L: $C = 64$, channel multiplier = $\{1,2,4,8\}$
\end{itemize}
where $\textit{C}$ is the channel number of the first hidden layers.

\subsection{Evaluation on the benchmarks}
\vspace{-1mm}
We compare our DiffUIR with several task-specific methods and six universal methods on five challenge image restoration tasks. The quantitative results are shown in Table~\ref{tab:all}. We systematically analyze our method compared with other universal models below: 
\par
\noindent\textbf{The necessity of universal image restoration task.} We reimplement the widely known Restomer~\cite{Restoreformer} in the universal setting by adding a learning prompt for each task. The result presents that directly applying the task-specific model to the universal setting is not a prudent choice.  
\par
\noindent\textbf{Comparison with universal methods.} We achieve outstanding performance improvement in all the tasks by a large margin (\ie, improve PSNR by 0.36 dB, 0.95 dB, 1.85 dB, 1.5 dB and 1.3 dB in Deraining, Low-light, Desnowing, Dehazing and Debluring tasks respectively), showing that selective hourglass mapping is a great solution for universal image restoration tasks. Note that we only use 1/5 parameters and 1/10 computational cost compared with other universal methods but perform outstanding performance. 
\par
\noindent\textbf{Comparison under different parameters.} To make a fair comparison, we provide several light versions of our DiffUIR. It could be seen that in each fair parameters level, our model outperforms the other universal models by a large margin. Notably, our DiffUIR-T comprises only 0.89M parameters, 1.72G computational cost with an outstanding performance. The great performance of different versions proves the scalability of the proposed DiffUIR.
\par
\noindent\textbf{Comparison with task-specific methods.} We even achieve comparable performance compared with task-specific methods in five image restoration tasks. The phenomenon proves that different tasks contain shared information which could accelerate single-task learning. 
\par
\noindent\textbf{Visual comparison.} The qualitative result is shown in Fig.~\ref{fig_light} to \ref{fig_blur}, and for more results please refer to \textit{Appendix} D. It could be seen that our DiffUIR generates more steady results in all the image restoration tasks compared with other task-specific (Restomer~\cite{Restoreformer}) and universal methods.

\begin{table*}[t]
    \centering
    \caption{Ablation study of each component in five image restoration tasks. EC, IC, SDT present explicit condition, implicit condition and shared distribution term respectively. We gradually remove the components to validate the synergistic effect.}
    \vspace{-3mm}
    \resizebox{2.0\columnwidth}{!}{
        \begin{tabular}{cc|c|c|cc|cc|cc|cc|cc}
        \toprule[0.12em]
        \multicolumn{2}{c|}{\multirow{2}{*}{\textbf{Method}}} & \multicolumn{1}{c|}{\multirow{2}{*}{\textbf{Condition level}}} &
        \multicolumn{1}{c|}{\multirow{2}{*}{\textbf{Shared level}}} & \multicolumn{2}{c|}{\textbf{Deraining} $(5sets)$} & \multicolumn{2}{c|}{\textbf{Enhancement}} & \multicolumn{2}{c|}{\textbf{Desnowing} $(2sets)$} & \multicolumn{2}{c|}{\textbf{Dehazing}} & \multicolumn{2}{c}{\textbf{Deblurring}} \\
        \multicolumn{2}{c|} {} & {} & {} & PSNR $\uparrow$ & SSIM $\uparrow$ & PSNR $\uparrow$ & SSIM $\uparrow$ & PSNR $\uparrow$ & SSIM $\uparrow$ & PSNR $\uparrow$ & SSIM $\uparrow$ & PSNR $\uparrow$ & SSIM $\uparrow$  \\
        \midrule[0.12em]
        \multicolumn{2}{c|}{DiffUIR} & \textbf{Strong} & \textbf{Strong} & \textbf{31.03} & \textbf{0.904} & \textbf{25.12} & \textbf{0.907} & \textbf{32.65} & \textbf{0.927} & \textbf{32.94} &\textbf{0.956} & \textbf{29.17} & \textbf{0.864} \\
        \midrule[0.1em]
        \multicolumn{2}{c|}{-SDT} & Strong & Weak & 30.34 & 0.898 & 22.99 & 0.898 & 32.34 & 0.923 & 32.47 & 0.945 & 28.50 & 0.850 \\
        \multicolumn{2}{c|}{-EC \& SDT} & Weak & Strong & 30.63 & 0.900 & 24.28 & 0.906 & 32.40 & 0.925 & 32.35 & 0.955 & 28.62 & 0.853 \\
        \multicolumn{2}{c|}{-IC \& SDT} & Weak & Weak & 22.93 & 0.703 & 20.93 & 0.546 & 23.96 & 0.773 & 24.26 & 0.889 & 27.08 & 0.814 \\
        \bottomrule[0.1em]
    \end{tabular}{}
    }
    \vspace{-2mm}
    \label{tab:uni_compare}
\end{table*}

\begin{table*}[h!]
    \centering
    \caption{Ablation study on shared distribution coefficient in five image restoration tasks.}
    \vspace{-3mm}
    \resizebox{1.8\columnwidth}{!}{
        \begin{tabular}{cc|cc|cc|cc|cc|cc}
        \toprule[0.12em]
        \multicolumn{2}{c|}{\multirow{2}{*}{\textbf{Uni ratio}}} &
        \multicolumn{2}{c|}{\textbf{Deraining} $(5sets)$} & \multicolumn{2}{c|}{\textbf{Enhancement}} & \multicolumn{2}{c|}{\textbf{Desnowing} $(2sets)$} & \multicolumn{2}{c|}{\textbf{Dehazing}} & \multicolumn{2}{c}{\textbf{Deblurring}} \\
        \multicolumn{2}{c|}{}  & PSNR $\uparrow$ & SSIM $\uparrow$ & PSNR $\uparrow$ & SSIM $\uparrow$ & PSNR $\uparrow$ & SSIM $\uparrow$ & PSNR $\uparrow$ & SSIM $\uparrow$ & PSNR $\uparrow$ & SSIM $\uparrow$ \\
        \midrule[0.12em]
        \multicolumn{2}{c|}{0.1} & 30.70 & 0.901 & 24.64 & 0.901 & 32.44 & 0.925 & 32.66 & 0.949 & 28.95 & 0.859 \\
        \multicolumn{2}{c|}{0.25} & 30.89 & 0.903 & 23.66 & 0.904 & 32.59 & 0.927 & 32.75 & 0.953 & 29.05 & 0.863 \\
        \multicolumn{2}{c|}{0.5} & 31.01 & 0.903 & 24.96 & 0.906 & 32.62 & 0.927 & 32.87 & 0.955 & 29.00 & 0.862 \\
        \multicolumn{2}{c|}{0.75} & 30.77 & 0.901 & 24.32 & 0.907 & 32.56 & 0.927 & 32.90 & 0.956 & 29.14 & 0.864 \\
        \multicolumn{2}{c|}{0.9} & \textbf{31.03} & \textbf{0.904} & \textbf{25.12} & \textbf{0.907} & \textbf{32.65} & \textbf{0.927} & \textbf{32.94} & \textbf{0.956} & \textbf{29.17} & \textbf{0.864} \\
        \multicolumn{2}{c|}{1} & 30.54 & 0.898 & 24.37 & 0.906 & 32.46 & 0.926 & 32.70 & 0.953 & 28.84 & 0.858\\
        \bottomrule[0.1em]
    \end{tabular}{}
    }
    \vspace{-4mm}
    \label{tab:ablate_uni}
\end{table*}

\subsection{Ablation Study}
\vspace{-1mm}
We conduct ablation study on all the tasks. We first evaluate the effectiveness of the strong condition guidance and shared distribution mapping, and validate that achieving both components is necessary; then we show the effectiveness of SDT in achieving synergistic effect, and show the impact of the batch size for our universal model learning.
\par
\noindent\textbf{The synergistic effect analysis.} We evaluate the effectiveness of the proposed shared distribution mapping and strong condition guidance in Table~\ref{tab:uni_compare}. We name the condition injected into the diffusion algorithm as explicit condition, the condition concatenates with the diffusing target as implicit condition. When erasing the SDT only, the model degrades to multi-partite mapping and the different distributions are not shared. The performance is impacted as the model is forced to learn different distributions separately, ignoring the shared information between different tasks; while erasing the explicit condition and SDT, the model degrades to the standard conditional diffusion model which weakens the condition level. Although it achieves shared distribution mapping, due to the mediated condition constraint, the performance drops dramatically; without the implicit condition and the SDT, the model neither provides strong condition guidance nor has shared distribution mapping and the image quality is impacted extremely. The experiments validate that equipping the model with the ability of strong condition guidance and shared distribution mapping simultaneously is significant for universal image restoration.
\par

\begin{figure}[t]
    \centering
    \includegraphics[width=1.\linewidth]{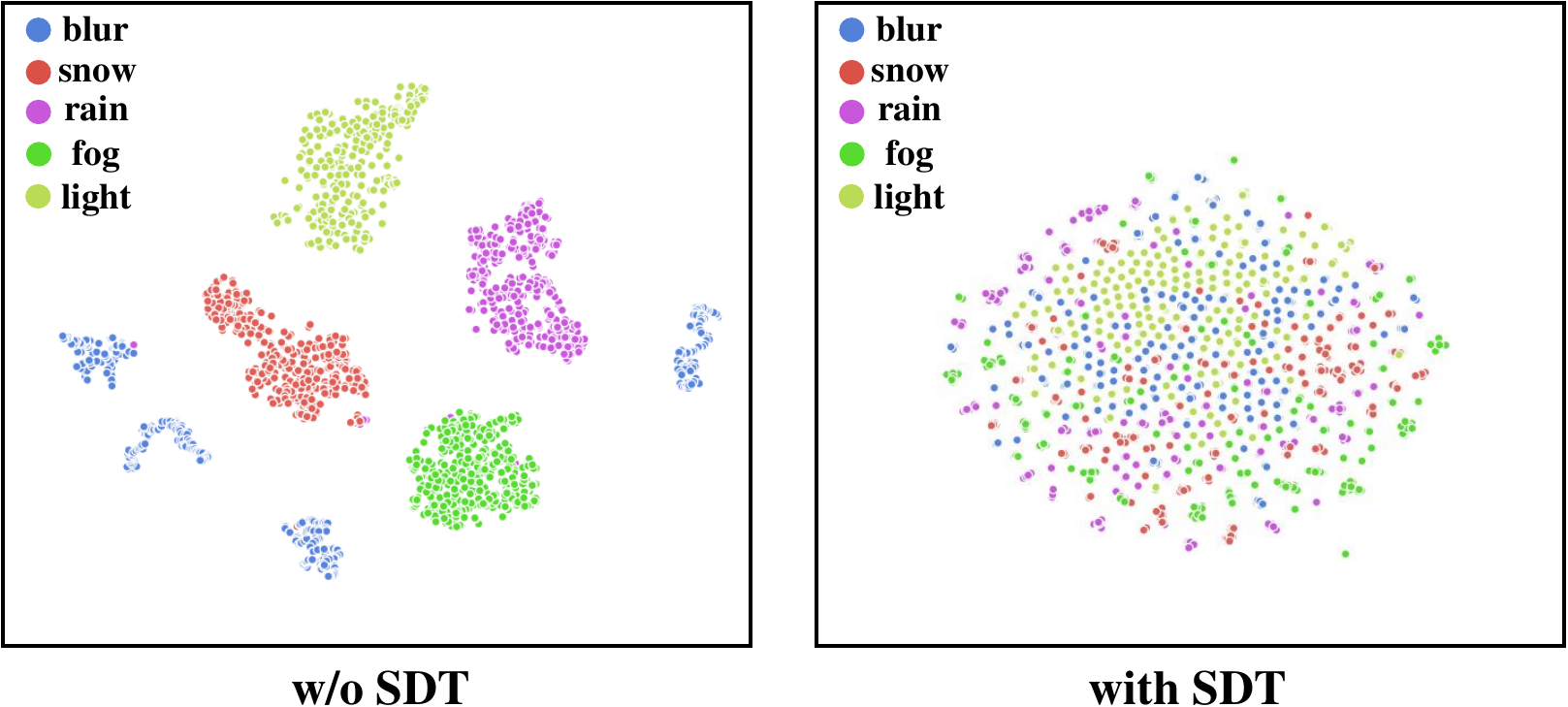}
    \vspace{-6mm}
    \caption{Distributions of the diffusing endpoint features of different image restoration tasks generated by w/o SDT and with SDT (value 0.9) visualized by t-SNE~\cite{tsne}. The diffusing endpoint features of different tasks are in different distributions without any interaction when without SDT. With the SDT, DiffUIR could achieve shared distribution mapping and strong condition guidance simultaneously. Zoom in for best view.}
    \label{fig:tsne}
    \vspace{-6mm}
\end{figure}

\begin{figure*}[t]
    \centering
    \includegraphics[width=1.\linewidth]{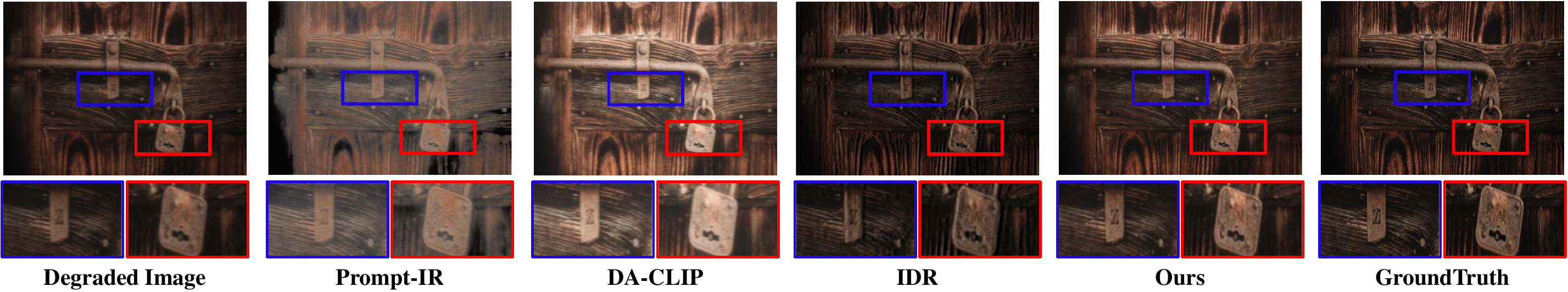}
    \vspace{-7mm}
    \caption{The visualization results of zero-shot generalization in real-world TOLED dataset. Zoom in for best view.}
    \vspace{-6mm}
    \label{fig:zero-shot}
\end{figure*}

\noindent\textbf{The effectiveness of SDT.} Table~\ref{tab:ablate_uni} presents that when shared distribution coefficient is under 0.9, the performance improves with the increase of the coefficient, as the larger the coefficient, the different distributions tend to converge towards the shared distribution. While up 0.9, the performance deteriorates with the coefficient increase because the weight of the condition in the diffusion algorithm tends to 0 and the model degrades to complete shared distribution mapping (\ie, pure Gaussian distribution). The experimental result proves that our SDT and impure Gaussian distribution are significant for universal image restoration learning.
\par
We further show the feature distributions of different tasks in Fig.~\ref{fig:tsne} to show the significance of the SDT. It could be seen that by formulating the SDT into the diffusion algorithm, the different degradation distributions map to a shared one, enabling the model to capture the shared information between different tasks. 
\par

\noindent\textbf{Impact of batch size.} We show the result in \textit{Appendix} D, the batch size is significant for our universal learning as the shared distribution mapping requires the model to sense all the tasks in the same batch.
\par

\noindent\textbf{More sampling steps.} We show the result in \textit{Appendix} D. Similar to other diffusion methods, DiffUIR will have a better performance with more sampling steps.
\par

\begin{table}[t]
    \centering
    \caption{Comparison under unknown tasks setting (under-display camera image restoration) on TOLED and POLED datasets.}
    \vspace{-3mm}
    \resizebox{1.\columnwidth}{!}{
        \begin{tabular}{ll|ccc|ccc}
        \toprule[0.15em]
        \multicolumn{2}{l|}{\multirow{2}{*}{\textbf{Method}}} &
        \multicolumn{3}{c|}{\textbf{TOLED}~\cite{udc}} & \multicolumn{3}{c}{\textbf{POLED}~\cite{udc}} \\
        \multicolumn{2}{c|}{}  & PSNR $\uparrow$ & SSIM $\uparrow$ & LPIPS $\downarrow$ & PSNR $\uparrow$ & SSIM $\uparrow$ & LPIPS $\downarrow$  \\
        \midrule[0.15em]
        \multicolumn{8}{l}{\textbf{Task-specific Method}}\\
        \midrule[0.1em]
        \multicolumn{2}{l|}{NAFNet~\cite{nafnet}} & \blue{26.89} & \blue{0.774} & \blue{0.346} & 10.83 & 0.416 & 0.794 \\
        \multicolumn{2}{l|}{HINet~\cite{hinet}} & 13.84 & 0.559 & 0.448 & 11.52 & \blue{0.436} & 0.831 \\
        \multicolumn{2}{l|}{MPRNet~\cite{mprnet}} & 24.69 & 0.707 & 0.347 & 8.34 & 0.365 & 0.798 \\
        \multicolumn{2}{l|}{DGUNet~\cite{dgunet}} & 19.67 & 0.627 & 0.384 & 8.88 & 0.391 & 0.810 \\
        \multicolumn{2}{l|}{MIRNetV2~\cite{mirnet}} & 21.86 & 0.620 & 0.408 & 10.27 & 0.425 & 0.722\\
        \multicolumn{2}{l|}{SwinIR~\cite{swinir}} & 17.72 & 0.661 & 0.419 & 6.89 & 0.301 & 0.852 \\
        \multicolumn{2}{l|}{RDDM~\cite{RDDM}$\dagger$} & 23.48 & 0.639 & 0.383 & \blue{15.58} & 0.398 & \blue{0.544} \\
        \multicolumn{2}{l|}{Restomer~\cite{Restoreformer}} & 20.98 & 0.632 & 0.360 & 9.04 & 0.399 & 0.742 \\
        \midrule[0.1em]
        \multicolumn{8}{l}{\textbf{Universal Method}}\\
        \midrule[0.1em]
        \multicolumn{2}{l|}{DL~\cite{dl}} & 21.23 & 0.656 & 0.434 & 13.92 & 0.449 & 0.756 \\
        \multicolumn{2}{l|}{Transweather~\cite{transweather}} & 25.02 & 0.718 & 0.356 & 10.46 & 0.422 & 0.760 \\
        \multicolumn{2}{l|}{TAPE~\cite{tape}} & 17.61 & 0.583 & 0.520 & 7.90 & 0.219 & 0.799 \\
        \multicolumn{2}{l|}{AirNet~\cite{airnet}} & 14.58 & 0.609 & 0.445 & 7.53 & 0.350 & 0.820 \\
        \multicolumn{2}{l|}{IDR~\cite{IDR}} & 27.91 & 0.795 & 0.312 & \red{16.71} & 0.497 & 0.716 \\
        \multicolumn{2}{l|}{Prompt-IR~\cite{promptir}$\dagger$} & 16.70 & 0.688 & 0.422 & 13.16 & \red{0.583} & 0.619 \\
        \multicolumn{2}{l|}{DA-CLIP~\cite{daclip}$\dagger$} & 15.74 & 0.606 & 0.472 & 14.91 & 0.475 & 0.739 \\
        \multicolumn{2}{l|}{\textbf{DiffUIR(ours)}}  & \red{29.55} & \red{0.887} & \red{0.281} & 15.62 & 0.424 & \red{0.505}  \\
        \bottomrule[0.1em]
    \end{tabular}{}
    }
    \vspace{-5mm}
    \label{tab:unknown}
\end{table}

\vspace{-1mm}
\subsection{Zero-shot generalization in real-world scenes}
\vspace{-1mm}
To evaluate the generalization ability of our DiffUIR, we do zero-shot known task generalization and unknown task generalization in real-world scenes respectively, where the difference lies in whether we know the degradation types. As our DiffUIR naturally owns the ability to handle different degradation types at once by selective hourglass mapping, we show excellent performance on both settings. 
\par
\noindent\textbf{Known task generalization.} As real-world datasets mainly have no ground truth, we use the perceptual evaluation metrics followed~\cite{WeatherDiff, npe_loe, IDR}. Table~\ref{tab:known} shows that our DiffUIR outperforms other universal models in various benchmarks. In each degradation task, we achieve comparable or even better results with task-specific methods.
\par
\noindent\textbf{Unknown task generalization.} The data~\cite{udc} is obtained by under-display camera in 1024$\times$2048 resolution with various degradation types, which meets the real-world scene. The SSIM and PSNR are computed in all channels. As each image combines several degradation types, task-specific models fail in nature and as existing universal methods mainly utilize the task-specific prompt to prompt the model, the mixing degradation types confuse the model prediction. In contrast, benefiting from selective hourglass mapping, we achieve state-of-the-art performance compared with all of the methods as shown in Table~\ref{tab:unknown}. 
\par
\noindent\textbf{Visual comparison.} We show our visualization results compared with other universal methods in Fig.~\ref{fig:zero-shot}. Our method could handle low-light and blurry degradations at once benefiting from the shared distribution mapping. 
\vspace{-1mm}
\subsection{Analysis of the shared distribution learning}
\vspace{-1mm}
To validate that we share useful information between different tasks, we analyze it from the perspectives below.
\par
\noindent\textbf{Theory.} According to~\cite{sde}, adding noise will change different data from low-dimensional manifolds to high-dimensional manifolds, which learn the attributes of different data at once, validating the rationality of our method.
\par
\begin{figure}[t]
    \centering
    \includegraphics[width=1\linewidth]{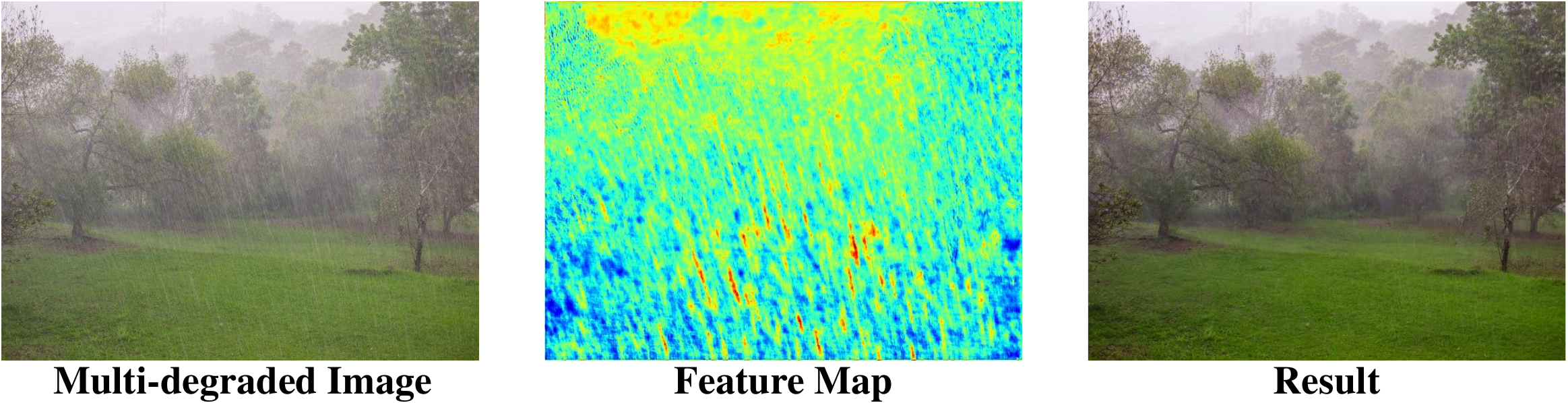}
    \vspace{-7mm}
    \caption{The visualization results of the attention of feature map during restoring multi-degraded images.}
    \vspace{-6mm}
    \label{fig:feature}
\end{figure}

\noindent\textbf{Visualization of the feature map.} In Fig.~\ref{fig:feature}, the attention of our feature map focuses on the region with rain and fog, validating that we share useful information.
\par

\begin{table}[t]
    \centering
    \caption{Comparison under known task generalization setting.}
    \vspace{-3mm}
    \setlength{\tabcolsep}{2.5pt}
    \resizebox{1.\columnwidth}{!}{
        \begin{tabular}{ll|cc|cc|cc|cc}
        \toprule[0.15em]
        \multicolumn{2}{l|}{\multirow{2}{*}{\textbf{Method}}} &
        \multicolumn{2}{c|}{\textbf{Deraining}} & \multicolumn{2}{c|}{\textbf{Enhancement}} & \multicolumn{2}{c|}{\textbf{Desnowing}} & \multicolumn{2}{c}{\textbf{Deblurring}} \\
        \multicolumn{2}{c|}{}  & NIQE$\downarrow$ & LOE$\downarrow$ & NIQE$\downarrow$ & LOE$\downarrow$ & NIQE$\downarrow$ & IL-NIQE$\downarrow$ & PSNR$\uparrow$ & SSIM$\uparrow$  \\
        \midrule[0.15em]
        \multicolumn{10}{l}{\textbf{Task-specific Method}}\\
        \midrule[0.1em]
        \multicolumn{2}{l|}{WeatherDiff~\cite{WeatherDiff}} & - & - & - & - & 2.96 & \blue{21.976} & - & - \\
        \multicolumn{2}{l|}{CLIP-LIT~\cite{clip-lit}} & - & - & 3.70 & 232.48 & - & - & - & - \\
        \multicolumn{2}{l|}{RDDM~\cite{RDDM}$\dagger$} & \blue{3.34} & 41.80 & \blue{3.57} & \blue{202.18} & \blue{2.76} & 22.261 & 30.74 & 0.894 \\
        \multicolumn{2}{l|}{Restomer~\cite{Restoreformer}} & 3.50 & \blue{30.32} & 3.80 & 351.61 & - & - & \blue{32.12} & \blue{0.926} \\
        \midrule[0.1em]
        \multicolumn{10}{l}{\textbf{Universal Method}}\\
        \midrule[0.1em]
        \multicolumn{2}{l|}{AirNet~\cite{airnet}$\dagger$} & 3.55 & 145.3 & 3.45 & 598.13 & 2.75 & 21.638 & 16.78 & 0.628 \\
        \multicolumn{2}{l|}{Prompt-IR~\cite{promptir}$\dagger$} & 3.52 & 28.53 & 3.31 & 255.13 & 2.79 & 23.000 & 22.48 & 0.77 \\
        \multicolumn{2}{l|}{DA-CLIP~\cite{daclip}$\dagger$} & 3.52 & 42.03 & 3.56 & 218.27 & \red{2.72} & \red{21.498} & 17.51 & 0.667 \\
        \multicolumn{2}{l|}{\textbf{DiffUIR(ours)}} & \red{3.38} & \red{24.82} & \red{3.14} & \red{193.40} & 2.74 & 22.426 & \red{30.63} & \red{0.890} \\
        \bottomrule[0.1em]
    \end{tabular}{}
    }
    \vspace{-3mm}
    \label{tab:known}
\end{table}

\vspace{-2mm}
\section{Conclusion}
\vspace{-1mm}
We present a selective hourglass mapping method based on conditional diffusion model, termed DiffUIR, that could achieve shared distribution mapping and strong condition guidance simultaneously for better universal image restoration learning. Specifically, we explicitly integrate the condition into the diffusion algorithm and extensively concatenate the condition to equip the model with strong condition guidance, generating accurate generation direction for diffusion model. More importantly, DiffUIR integrates a flexible shared distribution term (SDT) into the diffusion algorithm elegantly and naturally, which gradually maps different distributions into a shared one. The extensive experiments on five image restoration tasks, 22 benchmarks demonstrate that DiffUIR achieves state-of-the-art performance in the universal setting and zero-shot generalization setting compared with other universal methods.
\par
\noindent\textbf{Acknowledgments.} This work was supported partially by the National Key Research and Development Program of China (2023YFA1008503), NSFC(U21A20471), Guangdong NSF Project (No. 2023B1515040025, 2020B1515120085).

{
    \small
    \bibliographystyle{ieeenat_fullname}
    \bibliography{main}

\begin{thebibliography}{75}
\providecommand{\natexlab}[1]{#1}
\providecommand{\url}[1]{\texttt{#1}}
\expandafter\ifx\csname urlstyle\endcsname\relax
  \providecommand{\doi}[1]{doi: #1}\else
  \providecommand{\doi}{doi: \begingroup \urlstyle{rm}\Url}\fi

\bibitem[Avrahami et~al.(2022)Avrahami, Lischinski, and Fried]{blendededit}
Omri Avrahami, Dani Lischinski, and Ohad Fried.
\newblock Blended diffusion for text-driven editing of natural images.
\newblock In \emph{CVPR}, 2022.

\bibitem[Baranchuk et~al.(2021)Baranchuk, Rubachev, Voynov, Khrulkov, and Babenko]{baranchuk2021label}
Dmitry Baranchuk, Ivan Rubachev, Andrey Voynov, Valentin Khrulkov, and Artem Babenko.
\newblock Label-efficient semantic segmentation with diffusion models.
\newblock In \emph{ICLR}, 2021.

\bibitem[Biswas and Veloso(2011)]{2011robot}
Joydeep Biswas and Manuela Veloso.
\newblock Depth camera based localization and navigation for indoor mobile robots.
\newblock In \emph{RSS}, 2011.

\bibitem[Chen et~al.(2015)Chen, Seff, Kornhauser, and Xiao]{2015deepdriving}
Chenyi Chen, Ari Seff, Alain Kornhauser, and Jianxiong Xiao.
\newblock Deepdriving: Learning affordance for direct perception in autonomous driving.
\newblock In \emph{ICCV}, 2015.

\bibitem[Chen et~al.(2021)Chen, Lu, Zhang, Chu, and Chen]{hinet}
Liangyu Chen, Xin Lu, Jie Zhang, Xiaojie Chu, and Chengpeng Chen.
\newblock Hinet: Half instance normalization network for image restoration.
\newblock In \emph{CVPR}, 2021.

\bibitem[Chen et~al.(2022)Chen, Chu, Zhang, and Sun]{nafnet}
Liangyu Chen, Xiaojie Chu, Xiangyu Zhang, and Jian Sun.
\newblock Simple baselines for image restoration.
\newblock In \emph{ECCV}, 2022.

\bibitem[Chen et~al.(2023{\natexlab{a}})Chen, Sun, Song, and Luo]{diffusiondet}
Shoufa Chen, Peize Sun, Yibing Song, and Ping Luo.
\newblock Diffusiondet: Diffusion model for object detection.
\newblock In \emph{ICCV}, 2023{\natexlab{a}}.

\bibitem[Chen et~al.(2023{\natexlab{b}})Chen, Li, Li, and Pan]{drs}
Xiang Chen, Hao Li, Mingqiang Li, and Jinshan Pan.
\newblock Learning a sparse transformer network for effective image deraining.
\newblock In \emph{CVPR}, 2023{\natexlab{b}}.

\bibitem[Couairon et~al.(2022)Couairon, Verbeek, Schwenk, and Cord]{diffedit}
Guillaume Couairon, Jakob Verbeek, Holger Schwenk, and Matthieu Cord.
\newblock Diffedit: Diffusion-based semantic image editing with mask guidance.
\newblock In \emph{ICLR}, 2022.

\bibitem[Delbracio and Milanfar(2023)]{2023inver}
Mauricio Delbracio and Peyman Milanfar.
\newblock Inversion by direct iteration: An alternative to denoising diffusion for image restoration.
\newblock \emph{arXiv preprint arXiv:2303.11435}, 2023.

\bibitem[Fan et~al.(2019)Fan, Chen, Yuan, Hua, Yu, and Chen]{dl}
Qingnan Fan, Dongdong Chen, Lu Yuan, Gang Hua, Nenghai Yu, and Baoquan Chen.
\newblock A general decoupled learning framework for parameterized image operators.
\newblock \emph{TPAMI}, 2019.

\bibitem[He et~al.(2010)He, Sun, and Tang]{he2010single}
Kaiming He, Jian Sun, and Xiaoou Tang.
\newblock Single image haze removal using dark channel prior.
\newblock \emph{TPAMI}, 2010.

\bibitem[He et~al.(2016)He, Zhang, Ren, and Sun]{resnet}
Kaiming He, Xiangyu Zhang, Shaoqing Ren, and Jian Sun.
\newblock Deep residual learning for image recognition.
\newblock In \emph{CVPR}, 2016.

\bibitem[Ho et~al.(2020)Ho, Jain, and Abbeel]{ddpm}
Jonathan Ho, Ajay Jain, and Pieter Abbeel.
\newblock Denoising diffusion probabilistic models.
\newblock In \emph{NeurIPS}, 2020.

\bibitem[Huynh-Thu and Ghanbari(2008)]{psnr}
Quan Huynh-Thu and Mohammed Ghanbari.
\newblock Scope of validity of psnr in image/video quality assessment.
\newblock \emph{EL}, 2008.

\bibitem[Jiang et~al.(2020{\natexlab{a}})Jiang, Wang, Yi, Chen, Huang, Luo, Ma, and Jiang]{jiang2020multi}
Kui Jiang, Zhongyuan Wang, Peng Yi, Chen Chen, Baojin Huang, Yimin Luo, Jiayi Ma, and Junjun Jiang.
\newblock Multi-scale progressive fusion network for single image deraining.
\newblock In \emph{CVPR}, 2020{\natexlab{a}}.

\bibitem[Jiang et~al.(2020{\natexlab{b}})Jiang, Wang, Yi, Chen, Huang, Luo, Ma, and Jiang]{mspfn}
Kui Jiang, Zhongyuan Wang, Peng Yi, Chen Chen, Baojin Huang, Yimin Luo, Jiayi Ma, and Junjun Jiang.
\newblock Multi-scale progressive fusion network for single image deraining.
\newblock In \emph{CVPR}, 2020{\natexlab{b}}.

\bibitem[Kingma and Ba(2015)]{kingma2014adam}
Diederik~P Kingma and Jimmy Ba.
\newblock Adam: A method for stochastic optimization.
\newblock In \emph{ICLR}, 2015.

\bibitem[Kingma and Welling(2014)]{vae}
Diederik~P Kingma and Max Welling.
\newblock Auto-encoding variational bayes.
\newblock In \emph{ICLR}, 2014.

\bibitem[Kingma et~al.(2019)Kingma, Welling, et~al.]{inro_vae}
Diederik~P Kingma, Max Welling, et~al.
\newblock An introduction to variational autoencoders.
\newblock \emph{FTML}, 2019.

\bibitem[Lee et~al.(2012)Lee, Lee, and Kim]{dice}
Chulwoo Lee, Chul Lee, and Chang-Su Kim.
\newblock Contrast enhancement based on layered difference representation.
\newblock In \emph{ICIP}, 2012.

\bibitem[Li et~al.(2018)Li, Ren, Fu, Tao, Feng, Zeng, and Wang]{reside}
Boyi Li, Wenqi Ren, Dengpan Fu, Dacheng Tao, Dan Feng, Wenjun Zeng, and Zhangyang Wang.
\newblock Benchmarking single-image dehazing and beyond.
\newblock \emph{TIP}, 2018.

\bibitem[Li et~al.(2022)Li, Liu, Hu, Wu, Lv, and Peng]{airnet}
Boyun Li, Xiao Liu, Peng Hu, Zhongqin Wu, Jiancheng Lv, and Xi Peng.
\newblock All-in-one image restoration for unknown corruption.
\newblock In \emph{CVPR}, 2022.

\bibitem[Li et~al.(2020)Li, Tan, and Cheong]{multi-encoder}
Ruoteng Li, Robby~T Tan, and Loong-Fah Cheong.
\newblock All in one bad weather removal using architectural search.
\newblock In \emph{CVPR}, 2020.

\bibitem[Liang et~al.(2021)Liang, Cao, Sun, Zhang, Van~Gool, and Timofte]{swinir}
Jingyun Liang, Jiezhang Cao, Guolei Sun, Kai Zhang, Luc Van~Gool, and Radu Timofte.
\newblock Swinir: Image restoration using swin transformer.
\newblock In \emph{ICCV}, 2021.

\bibitem[Liang et~al.(2023)Liang, Li, Zhou, Feng, and Loy]{clip-lit}
Zhexin Liang, Chongyi Li, Shangchen Zhou, Ruicheng Feng, and Chen~Change Loy.
\newblock Iterative prompt learning for unsupervised backlit image enhancement.
\newblock In \emph{ICCV}, 2023.

\bibitem[Liu et~al.(2023{\natexlab{a}})Liu, Vahdat, Huang, Theodorou, Nie, and Anandkumar]{bridge}
Guan-Horng Liu, Arash Vahdat, De-An Huang, Evangelos~A Theodorou, Weili Nie, and Anima Anandkumar.
\newblock I$^2$ sb: Image-to-image schr{\"o}dinger bridge.
\newblock \emph{arXiv preprint arXiv:2302.05872}, 2023{\natexlab{a}}.

\bibitem[Liu et~al.(2023{\natexlab{b}})Liu, Wang, Fan, Wang, Tang, and Qu]{RDDM}
Jiawei Liu, Qiang Wang, Huijie Fan, Yinong Wang, Yandong Tang, and Liangqiong Qu.
\newblock Residual denoising diffusion models.
\newblock \emph{arXiv preprint arXiv:2308.13712}, 2023{\natexlab{b}}.

\bibitem[Liu et~al.(2022)Liu, Xie, Zhang, Yuan, Chen, Zhou, Li, and Tian]{tape}
Lin Liu, Lingxi Xie, Xiaopeng Zhang, Shanxin Yuan, Xiangyu Chen, Wengang Zhou, Houqiang Li, and Qi Tian.
\newblock Tape: Task-agnostic prior embedding for image restoration.
\newblock In \emph{ECCV}, 2022.

\bibitem[Liu et~al.(2018)Liu, Jaw, Huang, and Hwang]{snow100}
Yun-Fu Liu, Da-Wei Jaw, Shih-Chia Huang, and Jenq-Neng Hwang.
\newblock Desnownet: Context-aware deep network for snow removal.
\newblock \emph{TIP}, 2018.

\bibitem[Luo et~al.(2023{\natexlab{a}})Luo, Gustafsson, Zhao, Sj{\"o}lund, and Sch{\"o}n]{daclip}
Ziwei Luo, Fredrik~K Gustafsson, Zheng Zhao, Jens Sj{\"o}lund, and Thomas~B Sch{\"o}n.
\newblock Controlling vision-language models for universal image restoration.
\newblock \emph{arXiv preprint arXiv:2310.01018}, 2023{\natexlab{a}}.

\bibitem[Luo et~al.(2023{\natexlab{b}})Luo, Gustafsson, Zhao, Sj{\"o}lund, and Sch{\"o}n]{ir_sde}
Ziwei Luo, Fredrik~K Gustafsson, Zheng Zhao, Jens Sj{\"o}lund, and Thomas~B Sch{\"o}n.
\newblock Image restoration with mean-reverting stochastic differential equations.
\newblock \emph{ICML}, 2023{\natexlab{b}}.

\bibitem[Ma et~al.(2023)Ma, Cheng, Wang, Zhang, Wang, and Zhang]{Prores}
Jiaqi Ma, Tianheng Cheng, Guoli Wang, Qian Zhang, Xinggang Wang, and Lefei Zhang.
\newblock Prores: Exploring degradation-aware visual prompt for universal image restoration.
\newblock \emph{arXiv preprint arXiv:2306.13653}, 2023.

\bibitem[Ma et~al.(2015)Ma, Zeng, and Wang]{mef}
Kede Ma, Kai Zeng, and Zhou Wang.
\newblock Perceptual quality assessment for multi-exposure image fusion.
\newblock \emph{TIP}, 2015.

\bibitem[Mittal et~al.(2012)Mittal, Soundararajan, and Bovik]{niqe}
Anish Mittal, Rajiv Soundararajan, and Alan~C Bovik.
\newblock Making a “completely blind” image quality analyzer.
\newblock \emph{SPL}, 2012.

\bibitem[Mou et~al.(2022)Mou, Wang, and Zhang]{dgunet}
Chong Mou, Qian Wang, and Jian Zhang.
\newblock Deep generalized unfolding networks for image restoration.
\newblock In \emph{CVPR}, 2022.

\bibitem[Nah et~al.(2017)Nah, Hyun~Kim, and Mu~Lee]{gopro}
Seungjun Nah, Tae Hyun~Kim, and Kyoung Mu~Lee.
\newblock Deep multi-scale convolutional neural network for dynamic scene deblurring.
\newblock In \emph{CVPR}, 2017.

\bibitem[{\"O}zdenizci and Legenstein(2023)]{WeatherDiff}
Ozan {\"O}zdenizci and Robert Legenstein.
\newblock Restoring vision in adverse weather conditions with patch-based denoising diffusion models.
\newblock \emph{TPAMI}, 2023.

\bibitem[Paszke et~al.(2019)Paszke, Gross, Massa, Lerer, Bradbury, Chanan, Killeen, Lin, Gimelshein, Antiga, et~al.]{paszke2019pytorch}
Adam Paszke, Sam Gross, Francisco Massa, Adam Lerer, James Bradbury, Gregory Chanan, Trevor Killeen, Zeming Lin, Natalia Gimelshein, Luca Antiga, et~al.
\newblock Pytorch: An imperative style, high-performance deep learning library.
\newblock In \emph{NeurIPS}, 2019.

\bibitem[Potlapalli et~al.(2023)Potlapalli, Zamir, Khan, and Khan]{promptir}
Vaishnav Potlapalli, Syed~Waqas Zamir, Salman Khan, and Fahad~Shahbaz Khan.
\newblock Promptir: Prompting for all-in-one blind image restoration.
\newblock In \emph{NeurIPS}, 2023.

\bibitem[Purohit et~al.(2021)Purohit, Suin, Rajagopalan, and Boddeti]{purohit2021spatially}
Kuldeep Purohit, Maitreya Suin, AN Rajagopalan, and Vishnu~Naresh Boddeti.
\newblock Spatially-adaptive image restoration using distortion-guided networks.
\newblock In \emph{ICCV}, 2021.

\bibitem[Rim et~al.(2020)Rim, Lee, Won, and Cho]{realblur}
Jaesung Rim, Haeyun Lee, Jucheol Won, and Sunghyun Cho.
\newblock Real-world blur dataset for learning and benchmarking deblurring algorithms.
\newblock In \emph{ECCV}, 2020.

\bibitem[Rombach et~al.(2022)Rombach, Blattmann, Lorenz, Esser, and Ommer]{stable_diffusion}
Robin Rombach, Andreas Blattmann, Dominik Lorenz, Patrick Esser, and Bj{\"o}rn Ommer.
\newblock High-resolution image synthesis with latent diffusion models.
\newblock In \emph{CVPR}, 2022.

\bibitem[Ronneberger et~al.(2015)Ronneberger, Fischer, and Brox]{ronneberger2015unet}
Olaf Ronneberger, Philipp Fischer, and Thomas Brox.
\newblock U-net: Convolutional networks for biomedical image segmentation.
\newblock In \emph{MICCAI}, 2015.

\bibitem[Saharia et~al.(2022)Saharia, Chan, Chang, Lee, Ho, Salimans, Fleet, and Norouzi]{i2i}
Chitwan Saharia, William Chan, Huiwen Chang, Chris Lee, Jonathan Ho, Tim Salimans, David Fleet, and Mohammad Norouzi.
\newblock Palette: Image-to-image diffusion models.
\newblock In \emph{SIGGRAPH}, 2022.

\bibitem[Shao et~al.(2022)Shao, Zheng, Zhang, Sun, and Liu]{diffustereo}
Ruizhi Shao, Zerong Zheng, Hongwen Zhang, Jingxiang Sun, and Yebin Liu.
\newblock Diffustereo: High quality human reconstruction via diffusion-based stereo using sparse cameras.
\newblock In \emph{ECCV}, 2022.

\bibitem[Shen et~al.(2019)Shen, Wang, Lu, Shen, Ling, Xu, and Shao]{hide}
Ziyi Shen, Wenguan Wang, Xiankai Lu, Jianbing Shen, Haibin Ling, Tingfa Xu, and Ling Shao.
\newblock Human-aware motion deblurring.
\newblock In \emph{ICCV}, 2019.

\bibitem[Song et~al.(2020{\natexlab{a}})Song, Meng, and Ermon]{ddim}
Jiaming Song, Chenlin Meng, and Stefano Ermon.
\newblock Denoising diffusion implicit models.
\newblock In \emph{ICLR}, 2020{\natexlab{a}}.

\bibitem[Song and Ermon(2019)]{sde}
Yang Song and Stefano Ermon.
\newblock Generative modeling by estimating gradients of the data distribution.
\newblock In \emph{NeurIPS}, 2019.

\bibitem[Song et~al.(2020{\natexlab{b}})Song, Sohl-Dickstein, Kingma, Kumar, Ermon, and Poole]{score_image_generation}
Yang Song, Jascha Sohl-Dickstein, Diederik~P Kingma, Abhishek Kumar, Stefano Ermon, and Ben Poole.
\newblock Score-based generative modeling through stochastic differential equations.
\newblock In \emph{ICLR}, 2020{\natexlab{b}}.

\bibitem[Song et~al.(2023)Song, He, Qian, and Du]{DehezeFormer}
Yuda Song, Zhuqing He, Hui Qian, and Xin Du.
\newblock Vision transformers for single image dehazing.
\newblock \emph{TIP}, 2023.

\bibitem[Tu et~al.(2022)Tu, Talebi, Zhang, Yang, Milanfar, Bovik, and Li]{maxim}
Zhengzhong Tu, Hossein Talebi, Han Zhang, Feng Yang, Peyman Milanfar, Alan Bovik, and Yinxiao Li.
\newblock Maxim: Multi-axis mlp for image processing.
\newblock In \emph{CVPR}, 2022.

\bibitem[Valanarasu et~al.(2022)Valanarasu, Yasarla, and Patel]{transweather}
Jeya Maria~Jose Valanarasu, Rajeev Yasarla, and Vishal~M Patel.
\newblock Transweather: Transformer-based restoration of images degraded by adverse weather conditions.
\newblock In \emph{CVPR}, 2022.

\bibitem[Van~der Maaten and Hinton(2008)]{tsne}
Laurens Van~der Maaten and Geoffrey Hinton.
\newblock Visualizing data using t-sne.
\newblock \emph{JMLR}, 2008.

\bibitem[Wang et~al.(2023{\natexlab{a}})Wang, Wu, Guo, and Wang]{pdpp}
Hanlin Wang, Yilu Wu, Sheng Guo, and Limin Wang.
\newblock Pdpp: Projected diffusion for procedure planning in instructional videos.
\newblock In \emph{CVPR}, 2023{\natexlab{a}}.

\bibitem[Wang et~al.(2013)Wang, Zheng, Hu, and Li]{npe_loe}
Shuhang Wang, Jin Zheng, Hai-Miao Hu, and Bo Li.
\newblock Naturalness preserved enhancement algorithm for non-uniform illumination images.
\newblock \emph{TIP}, 2013.

\bibitem[Wang et~al.(2023{\natexlab{b}})Wang, Wang, Cao, Shen, and Huang]{Painter}
Xinlong Wang, Wen Wang, Yue Cao, Chunhua Shen, and Tiejun Huang.
\newblock Images speak in images: A generalist painter for in-context visual learning.
\newblock In \emph{CVPR}, 2023{\natexlab{b}}.

\bibitem[Wang et~al.(2022{\natexlab{a}})Wang, Yu, and Zhang]{ddnm}
Yinhuai Wang, Jiwen Yu, and Jian Zhang.
\newblock Zero-shot image restoration using denoising diffusion null-space model.
\newblock In \emph{ICLR}, 2022{\natexlab{a}}.

\bibitem[Wang et~al.(2004)Wang, Bovik, Sheikh, and Simoncelli]{ssim}
Zhou Wang, Alan~C Bovik, Hamid~R Sheikh, and Eero~P Simoncelli.
\newblock Image quality assessment: from error visibility to structural similarity.
\newblock \emph{TIP}, 2004.

\bibitem[Wang et~al.(2022{\natexlab{b}})Wang, Zhang, Chen, Wang, and Luo]{Restoreformer}
Zhouxia Wang, Jiawei Zhang, Runjian Chen, Wenping Wang, and Ping Luo.
\newblock Restoreformer: High-quality blind face restoration from undegraded key-value pairs.
\newblock In \emph{CVPR}, 2022{\natexlab{b}}.

\bibitem[Wei et~al.(2018)Wei, Wang, Yang, and Liu]{lol}
Chen Wei, Wenjing Wang, Wenhan Yang, and Jiaying Liu.
\newblock Deep retinex decomposition for low-light enhancement.
\newblock In \emph{BMVC}, 2018.

\bibitem[Wei et~al.(2023)Wei, Shen, Wang, Xie, and Wang]{raindiffusion}
Mingqiang Wei, Yiyang Shen, Yongzhen Wang, Haoran Xie, and Fu~Lee Wang.
\newblock Raindiffusion: When unsupervised learning meets diffusion models for real-world image deraining.
\newblock \emph{arXiv preprint arXiv:2301.09430}, 2023.

\bibitem[Xie et~al.(2021)Xie, Wang, Dong, Qi, and Shan]{single_learn}
Liangbin Xie, Xintao Wang, Chao Dong, Zhongang Qi, and Ying Shan.
\newblock Finding discriminative filters for specific degradations in blind super-resolution.
\newblock In \emph{NeurIPS}, 2021.

\bibitem[Xu et~al.(2023)Xu, Liu, Vahdat, Byeon, Wang, and De~Mello]{panoptic}
Jiarui Xu, Sifei Liu, Arash Vahdat, Wonmin Byeon, Xiaolong Wang, and Shalini De~Mello.
\newblock Open-vocabulary panoptic segmentation with text-to-image diffusion models.
\newblock In \emph{CVPR}, 2023.

\bibitem[Yang et~al.(2023)Yang, Ding, Wu, Li, and Zhang]{yangshuzhou}
Shuzhou Yang, Moxuan Ding, Yanmin Wu, Zihan Li, and Jian Zhang.
\newblock Implicit neural representation for cooperative low-light image enhancement.
\newblock In \emph{ICCV}, 2023.

\bibitem[Yang et~al.(2017)Yang, Tan, Feng, Liu, Guo, and Yan]{rain100}
Wenhan Yang, Robby~T Tan, Jiashi Feng, Jiaying Liu, Zongming Guo, and Shuicheng Yan.
\newblock Deep joint rain detection and removal from a single image.
\newblock In \emph{CVPR}, 2017.

\bibitem[Yu et~al.(2023)Yu, Wang, Zhao, Ghanem, and Zhang]{freedom}
Jiwen Yu, Yinhuai Wang, Chen Zhao, Bernard Ghanem, and Jian Zhang.
\newblock Freedom: Training-free energy-guided conditional diffusion model.
\newblock \emph{ICCV}, 2023.

\bibitem[Zamir et~al.(2021{\natexlab{a}})Zamir, Arora, Khan, Hayat, Khan, Yang, and Shao]{mprnet}
Syed~Waqas Zamir, Aditya Arora, Salman Khan, Munawar Hayat, Fahad~Shahbaz Khan, Ming-Hsuan Yang, and Ling Shao.
\newblock Multi-stage progressive image restoration.
\newblock In \emph{CVPR}, 2021{\natexlab{a}}.

\bibitem[Zamir et~al.(2021{\natexlab{b}})Zamir, Arora, Khan, Hayat, Khan, Yang, and Shao]{zamir2021multi}
Syed~Waqas Zamir, Aditya Arora, Salman Khan, Munawar Hayat, Fahad~Shahbaz Khan, Ming-Hsuan Yang, and Ling Shao.
\newblock Multi-stage progressive image restoration.
\newblock In \emph{CVPR}, 2021{\natexlab{b}}.

\bibitem[Zamir et~al.(2022)Zamir, Arora, Khan, Hayat, Khan, Yang, and Shao]{mirnet}
Syed~Waqas Zamir, Aditya Arora, Salman Khan, Munawar Hayat, Fahad~Shahbaz Khan, Ming-Hsuan Yang, and Ling Shao.
\newblock Learning enriched features for fast image restoration and enhancement.
\newblock \emph{TPAMI}, 2022.

\bibitem[Zhang et~al.(2023)Zhang, Huang, Yao, Yang, Yu, Zhou, and Zhao]{IDR}
Jinghao Zhang, Jie Huang, Mingde Yao, Zizheng Yang, Hu Yu, Man Zhou, and Feng Zhao.
\newblock Ingredient-oriented multi-degradation learning for image restoration.
\newblock In \emph{CVPR}, 2023.

\bibitem[Zhang et~al.(2015)Zhang, Zhang, and Bovik]{ilniqe}
Lin Zhang, Lei Zhang, and Alan~C Bovik.
\newblock A feature-enriched completely blind image quality evaluator.
\newblock \emph{TIP}, 2015.

\bibitem[Zhang et~al.(2018)Zhang, Isola, Efros, Shechtman, and Wang]{lpips}
Richard Zhang, Phillip Isola, Alexei~A Efros, Eli Shechtman, and Oliver Wang.
\newblock The unreasonable effectiveness of deep features as a perceptual metric.
\newblock In \emph{CVPR}, 2018.

\bibitem[Zheng et~al.(2023)Zheng, Wu, Liu, Meng, and Zheng]{diffuvolume}
Dian Zheng, Xiao-Ming Wu, Zuhao Liu, Jingke Meng, and Wei-shi Zheng.
\newblock Diffuvolume: Diffusion model for volume based stereo matching.
\newblock \emph{arXiv preprint arXiv:2308.15989}, 2023.

\bibitem[Zhou et~al.(2021)Zhou, Ren, Emerton, Lim, and Large]{udc}
Yuqian Zhou, David Ren, Neil Emerton, Sehoon Lim, and Timothy Large.
\newblock Image restoration for under-display camera.
\newblock In \emph{CVPR}, 2021.

\end{thebibliography}
}


\end{document}